\documentclass[fleqn,10pt]{wlscirep}
\usepackage[utf8]{inputenc}
\usepackage[T1]{fontenc}
\usepackage{graphicx}%
\usepackage{subfigure}%
\usepackage{amsmath,amssymb,amsfonts}%
\usepackage{amsthm}%
\usepackage{amssymb}
\usepackage{mathtools}
\usepackage[makeroom]{cancel}
\usepackage{array}
\usepackage[title]{appendix}%
\usepackage{xcolor}%
\usepackage{textcomp}%
\usepackage{verbatim}%
\usepackage{manyfoot}%
\usepackage{booktabs}%
\usepackage{multirow}%
\usepackage{algorithm}%
\usepackage{algorithmicx}%
\usepackage{algpseudocode}%
\usepackage{listings}%
\usepackage{float}

\newtheorem{theorem}{Theorem}
\renewenvironment{proof}{{\bfseries Proof: }}{\qed}

\title{Time-Scale Separation in Q-Learning: Extending TD($\triangle$) for Action-Value Function Decomposition}

\author[1,2,3,*]{Mahammad Humayoo}


\begin{abstract}
Q-Learning is a fundamental off-policy reinforcement learning (RL) algorithm that has the objective of approximating action-value functions in order to learn optimal policies. Nonetheless, it has difficulties in reconciling bias with variance, particularly in the context of long-term rewards. This paper introduces Q($\Delta$)-Learning, an extension of TD($\Delta$) for the Q-Learning framework. TD($\Delta$) facilitates efficient learning over several time scales by breaking the Q($\Delta$)-function into distinct discount factors. This approach offers improved learning stability and scalability, especially for long-term tasks where discounting bias may impede convergence. Our methodology guarantees that each element of the Q($\Delta$)-function is acquired individually, facilitating expedited convergence on shorter time scales and enhancing the learning of extended time scales. We demonstrate through theoretical analysis and practical evaluations on standard benchmarks like Atari that Q($\Delta$)-Learning surpasses conventional Q-Learning and TD learning methods in both tabular and deep RL environments.

\end{abstract}
\begin{document}

\flushbottom
\maketitle
%
%
\thispagestyle{empty}


\section{Introduction}
\label{introduction}
In reinforcement learning (RL), the goal is to optimize an agent's actions to maximize cumulative rewards via interaction with an environment. Temporal difference (TD) learning techniques, such as Q-learning and SARSA, have shown effective in various tasks by enabling agents to estimate action-value functions that predict expected future rewards. Traditionally, these strategies employ a discount factor $0\leq\gamma<1$, where values nearing $\gamma=0$ prioritize short-term rewards over long-term ones, hence reducing the planning horizon and improving learning stability and efficiency\cite{sutton2018reinforcement}. Discount factors $\gamma<1$ often produce superior outcomes in the initial stages of learning, as evidenced by Prokhorov and Wunsch\cite{prokhorov1997adaptive}. In particular circumstances, such as long-horizon activities\cite{mnih2013playing,berner2019dota} where long-term planning is crucial—such as navigation or decision-making tasks with delayed rewards—this strategy may introduce bias, limiting the agent's ability to develop optimal long-horizon policies.

Q-Learning is a prevalent RL method that modifies the action-value function Q(s, a) in accordance with the Bellman equation. Despite their efficacy, conventional Q-Learning techniques frequently depend on a static discount factor $\gamma$, which may not be appropriate for all time scales\cite{watkins1992q}. A diminutive $\gamma$ may lead the agent to excessively prioritize immediate rewards\cite{tsitsiklis1996analysis}, whereas a large $\gamma$ amplifies the variance of value estimations\cite{peng1993convergence}, resulting in unstable learning\cite{sutton2018reinforcement}.

Recent research in TD learning has explored the segmentation of value functions across several time scales, allowing agents to simultaneously improve value estimates at various horizons\cite{romoff2019separating}. The TD($\Delta$) design improves scalability and performance in scenarios with long-term dependencies by deconstructing value functions into components that use diverse discount factors, known as TD($\Delta$). The adaptability of the TD($\Delta$) paradigm shows considerable promise in alleviating the bias-variance trade-offs intrinsic to TD learning.

Expanding on this concept, we introduce Q($\Delta$)-Learning, an extension of TD($\Delta$)\cite{romoff2019separating,de2018multi,szepesvari2022algorithms}, which decomposes the action-value function into delta components linked to various discount factors. TD($\Delta$) highlights the actor-critic framework and traditional TD learning, while Q($\Delta$)-Learning decomposes the state-action value function into many partial estimators across different discount factors, referred to as delta estimators ($W_{z}(s, a))$. These estimators approximate the difference $W_{z}(s, a) = Q_{\gamma_{z}}(s, a) \mathit{-} Q_{\gamma_{z -1}}(s, a)$ between action-value functions, therefore enhancing learning efficiency in contexts where actions influence long-term results. Q($\Delta$)-Learning optimizes a series of delta estimators, each linked to a certain discount factor, akin to TD($\Delta$). The segmentation facilitates the independent learning of each component, with lower discount factors resulting in quicker convergence, while higher discount factors bolster this foundation for enhanced long-horizon planning. This approach improves the adaptability and efficacy of Q-Learning by enabling the agent to learn over various time scales.

We present a straightforward method for finding intermediary $\gamma$ values, which frequently enhances performance in Q-learning without necessitating further hyperparameter optimization. Nevertheless, we demonstrate that this strategy permits further fine-tuning to attain enhanced performance improvements. Furthermore, our methodology aligns with adaptive $\gamma$ selection strategies\cite{xu2018meta}. We illustrate these benefits both theoretically and empirically, evidencing performance enhancements in a basic ring MDP—utilized by Kearns \& Singh\cite{kearns2000bias} for bias-variance analysis—by modifying the k-step returns in the Q-learning updates. Additionally, we demonstrate the integration of our method with techniques such as TD($\lambda$)\cite{sutton1984temporal} and Generalized Advantage Estimation (GAE)\cite{schulman2015high}, resulting in enhanced empirical performance in dense reward environments, including Atari games\cite{romoff2019separating}.

One incentive for expanding TD($\Delta$) to Q($\Delta$)-learning is that the Temporal Difference (TD($\Delta$)) learning framework addresses the restrictions of employing a singular discount factor in long-horizon tasks by decomposing the Q-function across several time scales. This multi-timescale learning method can alleviate the bias-variance tradeoff by allowing the agent to concurrently learn action-values for several discount factors. Extending TD($\Delta$) to Q-Learning enables the algorithm to partition the Q-function Q(s,a) into delta components over various time scales, hence enhancing its capacity to efficiently handle both short-term and long-term rewards.

This paper's principal contributions are as follows: (i) We enhance the TD($\Delta$) framework for Q-Learning by partitioning the Q-function Q(s,a) over various discount factors $\gamma_0, \gamma_1, ..., \gamma_Z$. (ii) The TD($\Delta$)-Q-Learning algorithm acquires distinct delta components for each discount factor, which are subsequently amalgamated to reconstruct the complete Q-function. (iii) This method improves the stability and efficiency of Q-Learning in long-horizon tasks by tackling the bias-variance trade-off and enabling the agent to learn across various time scales.

The subsequent sections of the paper are structured as follows: Section \ref{relatedwork} contains a discussion of related works. Section \ref{Background} presents the necessary background information. Sections \ref{method} and \ref{TA} illustrate the principles of Q($\Delta$)-learning and theoretical analysis, respectively. Section \ref{experiment} presents a comprehensive overview of the experiments conducted. Ultimately, we provide a conclusion in section \ref{conclusion}.

\section{Related Work}
\label{relatedwork}
In reinforcement learning (RL), optimising for long-horizon rewards is tough due to the complexity of learning with an undiscounted return. Temporal discounting is frequently employed to streamline this process; nevertheless, it may induce bias. Researchers\cite{romoff2019separating} have investigated techniques to separate value functions across various time scales by analysing disparities among value functions with lower discount factors, hence enhancing scalability and performance. This\cite{van2016deep} study presents enhancements to Q-Learning by mitigating overestimation bias, a concept particularly pertinent when adapting Q-Learning to TD($\Delta$). It examines the bias-variance tradeoff and provides justification for decomposing Q-values to more effectively address this tradeoff in long-horizon tasks. This study\cite{de2018multi} presents methodologies for multi-step RL, wherein learning transpires over many time scales (analogous to TD($\Delta$)). It provides theoretical understanding of how learning with various discount factors can stabilize Q-value updates.

Researchers\cite{kearns2000bias} offer an empirical study of the bias-variance trade-off in TD updates. It establishes limits for the error generated by employing TD learning, which directly facilitates the assessment of bias-variance tradeoffs in Q($\Delta$)-Learning. This study\cite{munos2003error} examines error boundaries and convergence assurances for RL algorithms. It substantiates the assertion that Q($\Delta$)-Learning inherits the convergence characteristics of Q-Learning while tackling the complexities posed by approximation policy iteration in the decomposition of Q-values along time-scales.

The authors\cite{schaul2015prioritized} of this study describe an addition to Q-Learning by the implementation of prioritized experience replay, enabling the algorithm to more effectively manage long-horizon dependencies. The text also examines how sampling tactics might diminish variation, hence reinforcing the necessity for multi-time-scale updates in Q($\Delta$)-Learning. This study\cite{sherstan2018generalizing} examines value estimation across several time-scales, similar to the TD($\Delta$) paradigm. It presents both theoretical and empirical findings that illustrate the advantages of multi-timescale learning, which aligns closely with the objectives of Q($\Delta$)-Learning. This work\cite{osband2016generalization} examines randomized value functions and generalization over time-scales, providing empirical evidence that substantiates the efficacy of Q-function decomposition techniques like Q($\Delta$)-Learning.

Recent study has focused on the exact determination of the discount factor\cite{prokhorov1997adaptive,franccois2015discount,xu2018meta,berner2019dota}. The authors\cite{xu2018meta} of this study present meta-gradient methods for the adaptive selection of discount factors in RL, highlighting the importance of balancing short-term and long-term rewards, which aligns with the reasoning for analyzing action-value functions across multiple discount factors in Q($\Delta$)-Learning.

Finally, another significant area of study relevant to our research indirectly includes hierarchical RL and the bias-variance tradeoff. The referenced studies\cite{Dietterich1999HierarchicalRL,Henderson2017OptionGANLJ,Hengst2002DiscoveringHI,Reynolds1999DecisionBP,Menache2002QCutD,Russell2003QDecompositionFR,Seijen2017HybridRA} introduce the notion of partitioning value functions into subtasks through hierarchical RL, similar to how TD($\Delta$) disaggregates value functions across different time-scales. This supports the claim that value function decomposition can improve learning efficiency and stability. This paper\cite{kearns2000bias} serves as a theoretical basis for understanding the bias-variance tradeoff involved in decomposing Q-values over different time-scales and sets limitations on the bias and variance of TD updates. This study\cite{Mnih2015HumanlevelCT} provides empirical evidence of the effectiveness of deep RL methods employing value function approximation in Atari-like settings. It also advocates for methods that reduce variance in long-horizon tasks.

\section{Background and notation}
\label{Background}
Analyze a fully observable Markov Decision Process (MDP) \cite{bellman1957markovian}, defined by the tuple $(\mathcal{S}, \mathcal{A}, \mathcal{P}, r)$, where $\mathcal{S}$ represents the state space, $\mathcal{A}$ indicates the action space, and $\mathcal{P}: \mathcal{S} \times \mathcal{A} \rightarrow \mathcal{S} \rightarrow [0, 1]$ signifies the transition probabilities linking state-action pairs to distributions across following states, while $r: \mathcal{S} \times \mathcal{A} \rightarrow \mathbb{R}$ represents the reward function. At each timestep $t$, the agent resides in state $s_t$, selects an action $a_t$, receives a reward $r_t = r(s_t, a_t)$, and transitions to the next state $s_{t+1} \sim \mathcal{P}(\cdot|s_t, a_t)$.

In a standard Markov Decision Process (MDP) paradigm, an agent aims to optimize the discounted return expressed as $Q_{\gamma}^{\pi}(s, a) = \mathbb{E}\left[\sum_{t=0}^{\infty} \gamma^{t} r_{t+1} \mid s_{t} = s, a_{t} = a\right]$, where $\gamma$ represents the discount factor and $\pi: \mathcal{S} \rightarrow \mathcal{A} \rightarrow [0, 1]$denotes the policy followed by the agent. The action-value function $Q_{\gamma}^{\pi}(s, a)$ is established as the fixed point of the Bellman operator $\mathcal{T} Q = r^{\pi} + \gamma \mathcal{P}^{\pi} Q$, where $r^{\pi}$ signifies the expected immediate reward and $\mathcal{P}^{\pi}$ denotes the transition probability operator linked to the policy $\pi$. For simplicity, we will omit the superscript $\pi$ for the remainder of the paper.

Through temporal difference (TD) learning\cite{sutton1984temporal}, the action-value estimate $\hat{Q}_{\gamma}$ can resemble the actual action-value function $Q_{\gamma}$. The one-step TD error $\delta^{\gamma}_{t} = r_{t+1} + \gamma \max_{a}\hat{Q}_{\gamma}(s_{t+1}, a) - \hat{Q}_{\gamma}(s_{t}, a_{t})$ is employed to update the action-value function based on the transition $(s_{t}, a_{t}, r_{t}, s_{t+1})$.

Q-Learning is an off-policy RL method that acquires the optimal action-value function $Q^*(s, a)$ by iterative updates of the Q-values, depending on observed rewards and the maximum estimated future reward. We define the Q-Learning update rule as follows:
\begin{align}
\label{Eq1}
Q(s_t, a_t) \leftarrow Q(s_t, a_t) + \alpha \left[ r_t + \gamma \max_a Q(s_{t+1}, a) - Q(s_t, a_t) \right]
\end{align}

where $\alpha$ indicates the learning rate, $\gamma$ specifies the discount factor, and $r_{t}$ represents the immediate reward. The algorithm is "off-policy" as the updates rely on the action that optimizes future rewards, irrespective of the action executed by the agent. Although Q-Learning can proficiently acquire the optimal policy, it encounters difficulties in contexts where long-term dependencies are crucial, mainly because of the fixed discount factor.

Conversely, for a comprehensive trajectory, we can utilize the discounted summation of one-step TD errors, commonly referred to as the $\lambda$-return\cite{sutton1984temporal} or, alternatively, the Generalized Advantage Estimator (GAE)\cite{Schulman2015HighDimensionalCC}. The GAE improves advantage estimates by balancing the trade-off between variance and bias through the parameters $\lambda$ and $\gamma$. The formula for the Generalized Advantage Estimator is typically stated as:
\begin{align}
\label{NGAE}
A(s_{t}, a_{t}) = \sum_{k=0}^{\infty}(\lambda \gamma)^{k}\delta_{t+k}^{\gamma}
\shortintertext{Where $\delta_{t+k}$ is the TD error at time t, computed as follows: $\delta_{t+k} = r_{t+k} + \gamma \quad max_{a}Q(s_{t+k+1}, a) - Q(s_{t+k}, a_{t+k})$}\nonumber
\end{align}
Loss function for Q-value estimate employing Generalized Advantage estimate (GAE). The loss function $\mathcal{L}(\theta)$ for approximating the Q-value function is defined as the mean squared error between the current Q-value estimate $Q(s_t, a_t;\theta)$ and the target Q-value modified by the advantage estimator. Consequently, the loss function can be succinctly expressed using $A(s_t, a_t)$ and $Q(s_t, a_t)$ as follows:
\begin{align}
\label{lossfun}
\mathcal{L}(\theta) = \mathbb{E} \bigg[\bigg(Q(s_t, a_t;\theta) - \bigg(Q(s_t, a_t) + A(s_t, a_t)\bigg)\bigg)^{2}\bigg]
\end{align}
As an off-policy method, Q-learning updates are determined by the maximum future Q-value. To implement Eq.\ref{ppoloss} in a Q-learning framework. In actor-critic architectures\cite{Sutton1999PolicyGM,Konda1999ActorCriticA, Mnih2016AsynchronousMF}, the action-value function is modified in accordance with Eq.\ref{lossfun}, while a stochastic parameterized policy (actor, $\pi_\omega(a|s)$) is derived from this value estimator via the advantage function, with the associated loss being.
\begin{align}
\label{ppoloss}
\mathcal{L}(\omega) = \mathbb{E} \bigg[-log\pi(a|s;\omega)A(s,a)\bigg]
\end{align}
Proximal Policy Optimization (PPO)\cite{schulman2017proximal}, an advancement of actor-critic approaches, restricts policy updates to a specified optimization region, known as a trust region, with a clipping objective that compares the current parameters, $\omega$, with the previous parameters, $\omega_{old}$:
\begin{align}
\label{ppoobj}
\mathcal{L}(\omega) = \mathbb{E}\bigg[min\bigg(\rho(\omega)A(s,a),\psi(\omega)A(s,a)\bigg)\bigg]
\end{align}
The likelihood ratio $\rho(\omega)$ indicates that Q-learning, being off-policy, does not depend directly on the policy for action selection but instead updates according to the greatest Q-value. Nevertheless, when employed within an actor-critic framework, the Q-learning agent may adopt a comparable policy ratio:
\begin{align}
\label{ppoclipp}
\rho(\omega) = \frac{max_{a}Q(s, a;\omega)}{Q(s, a;\omega_{old})}
\end{align}
where $\psi(\omega)= clip(\rho, 1-\epsilon, 1+\epsilon)$ denotes the clipped likelihood ratio, and $\epsilon < 1$ is a minor factor utilized to limit the update. The loss function for the policy update would be expressed as follows:
\begin{align}
\label{cliplossfun}
\mathcal{L}(\omega) = \mathbb{E}\bigg[min\bigg(\rho(\omega)A(s,a),clip(\rho(\omega), 1-\epsilon, 1+\epsilon)A(s,a)\bigg)\bigg]
\end{align}

\section{Methodology}
\label{method}
The TD($\Delta$) framework improves traditional temporal difference learning by segmenting action-value functions over many discount factors $\gamma_{0}, \gamma_{1}, ..., \gamma_{z}$. This decomposition allows the action-value function to be obtained as a summation of delta estimators, each reflecting the difference between action-value functions at different discount factors. The primary benefit of this strategy is its ability to control variance and bias in the learning process by focusing on smaller, more manageable time-scales.

\subsection{TD($\Delta$) Framework}
\label{TDF}
To compute $W_{z}$ for Q-learning as described in the study, it is crucial to understand how delta estimators ($W_{z}$) approximate the differences between action-value functions at successive discount factors. Let us outline this methodology for Q-learning in a systematic manner.\\
\textbf{Delta Estimators ($W_{z}$) and Action-Value Functions ($Q(s, a)$)}:
Delta estimators, $W_{z}$, denote the disparity between action-value functions with consecutive discount factors:
\begin{align}
\label{Eq2}
   W_{z}(s, a) = Q_{\gamma_{z}}(s, a) \mathit{-} Q_{\gamma_{z -1}}(s, a)
\end{align}

In this context, $Q_{\gamma_{z}}$ represents the action-value function with discount factor $\gamma_{z}$, but $Q_{\gamma_{z-1}}$ indicates the action-value function with the preceding discount factor, $\gamma_{z-1}$.
\subsection{Single-Step TD (Q($\Delta$)-Learning)}
\label{td-q-learning}
This section commences with an exposition of the delta estimators ($W_{z}$) for the action-value functions $Q$ employed in Q-learning. The extension of TD($\Delta$) to Q-learning is referred to as Q($\Delta$)-Learning. In Q($\Delta$)-learning, the update process focuses on maximizing the action-value at the subsequent state instead than adhering to a particular policy, as is the case in SARSA. The objective is to enhance the total reward by optimizing the action-value function over various time scales. The fundamental tenet of Q($\Delta$)-Learning entails partitioning the action-value function $Q(s, a)$ into a sequence of delta components $W_z(s, a)$, each linked to a distinct discount factor $\gamma_z$. The relationship among these delta components (i.e., delta function) is expressed as follows:
\begin{align}
\label{destimator1}
W_{z}(s, a) := Q_{\gamma_{z}}(s, a) \mathit{-} Q_{\gamma_{z -1}}(s, a)
\end{align}
where $\gamma_{0}, \gamma_{1}, \ldots, \gamma_{z}$ denote the discount factors across different time-scales, and define $W_{0}(s,a) := Q_{\gamma_{0}}(s, a)$. The action-value function $Q_{\gamma_z}(s, a)$ is defined as the aggregate of all W-components up to z:
\begin{align}
\label{act-val-fun}
Q_{\gamma_Z}(s, a) = \sum_{z=0}^{Z} W_z(s, a)
\end{align}
This collection of delta components yields the comprehensive action-value function $Q_{\gamma_Z}(s, a)$, which incorporates both short-term and long-term reward information across several time-scales. Consequently, Q($\Delta$)-Learning is more adept at managing long-horizon tasks, where the equilibrium between immediate and deferred rewards is essential.

In Q-learning, we select the action that optimizes the action-value at the subsequent state $s_{t+1}$ rather than following to a policy-determined action. The action-value function is modified via the TD error, with updates being off-policy and aimed at maximizing action values at each iteration:
\begin{align}
\label{QNorm}
Q(s_t, a_t) \leftarrow Q(s_t, a_t) + \alpha \left[ r_t + \gamma \quad max_{a}Q(s_{t+1}, a) - Q(s_t, a_t) \right]
\end{align}
The Q-learning update rule is applied separately to each delta component $W_z(s, a)$. The update for each component use the same structure as conventional Q-learning, but it is modified to include the delta function at each distinct time-scale. The update equation Eq.\ref{QNorm} for single-step TD Q-learning ($\Delta$) can be articulated with different time-scales by employing the principle of partitioning the update into several discount factors\cite{romoff2019separating} as demonstrated below:
\begin{align}
\label{QDelta}
W_z(s_t, a_t) = \mathbb{E} \left[ (\gamma_z - \gamma_{z-1})max_{a} Q_{\gamma_{z-1}}(s_{t+1}, a) + \gamma_z W_z(s_{t+1}, a) \right]
\end{align}
This Eq. combines both the delta decomposition over time-scales (with $\gamma_z$ and $\gamma_{z-1}$) and the greedy action selection of Q-learning (through the $\max_{a}$ term).\\
where $Q_{\gamma_z}(s_t, a_t)$ denotes the action-value function with a discount factor. The delta function for Q($\Delta$)-learning is denoted by $W_z(s_t, a_t)$, while $\gamma_z$ represents the discount factor for time-scale z. The original TD update is mirrored by the single-step TD Q($\Delta$)-learning, which is applied to action-value functions Q rather than state-value functions V. The Q-value function in standard Q-learning is represented by the single-step Bellman Eq. as follows:
\begin{align}
\label{BellmanEq}
Q_{\gamma_{z}}(s_{t}, a_{t}) = \mathbb{E}\left[r_{t} + \gamma_{z}max_{a}Q_{\gamma_{z}}(s_{t+1}, a)\right]
\end{align}
This Eq. indicates that the expected value of executing action $a_{t}$ in state $s_{t}$ is the immediate reward $r_{t}$, in addition to the discounted maximum future value $max_{a}Q(s_{t+1},a)$ from the subsequent state. Now, We propose to utilize a sequence of discount factors $\gamma_{0}, \gamma_{1}, \ldots, \gamma_{z}$ rather than a singular discount factor $\gamma$, and to decompose the action-value function into delta components. The delta function $W_{z}(s_{t}, a_{t})$ is defined as the difference between action-value functions at successive discount factors from Eq.\ref{destimator1}:
\begin{align}
\label{destimator2}
W_z(s_t, a_t) = Q_{\gamma_z}(s_t, a_t) - Q_{\gamma_{z-1}}(s_t, a_t)
\end{align}
In order to express $W_z$ in terms of these two action-value functions, we now replace $Q_{\gamma_z}$ and $Q_{\gamma_{z-1}}$ with the Bellman optimality Eq. as follows:
\begin{align}
W_z(s_t, a_t) = Q_{\gamma_z}(s_t, a_t) - Q_{\gamma_{z-1}}(s_t, a_t) \text{ From Eq. \ref{destimator2}}\\ \nonumber
W_z(s_t, a_t) = \mathbb{E}\left[r_t + \gamma_z max_{a}Q_{\gamma_z}(s_{t+1}, a)\right] - \mathbb{E}\left[r_t + \gamma_{z-1} max_{a}Q_{\gamma_{z-1}}(s_{t+1}, a)\right] \text{ From Eq. \ref{BellmanEq}}\\ \nonumber
\shortintertext{Simplify by removing the reward terms $r_t$, which appear in both formulations.}
W_z(s_t, a_t) = \mathbb{E}\left[\gamma_z max_{a}Q_{\gamma_z}(s_{t+1}, a) - \gamma_{z-1} max_{a}Q_{\gamma_{z-1}}(s_{t+1}, a)\right] \nonumber
\shortintertext{The expression can be more thoroughly analyzed by decomposing the terms. Employing the recursive relationship from Eq. \ref{destimator2}}
Q_{\gamma_z}(s_{t+1}, a) = W_z(s_{t+1}, a) + Q_{\gamma_{z-1}}(s_{t+1}, a) \nonumber
\shortintertext{Replace the definition of $W_z$ in the Eq. provided below.}
W_z(s_t, a_t) = \mathbb{E}\left[\gamma_z max_{a}(W_z(s_{t+1}, a) + max_{a}Q_{\gamma_{z-1}}(s_{t+1}, a)) - \gamma_{z-1} max_{a}Q_{\gamma_{z-1}}(s_{t+1}, a)\right] \nonumber
\shortintertext{Simplify the terminology:}
W_z(s_t, a_t) = \mathbb{E}\left[\gamma_z max_{a}W_z(s_{t+1}, a) + (\gamma_z - \gamma_{z-1}) max_{a} Q_{\gamma_{z-1}}(s_{t+1}, a)\right] \nonumber
\shortintertext{Combine the terms to get the final update Eq. for $W_z(s_t, a_t)$.}
W_z(s_t, a_t) = \mathbb{E}\left[(\gamma_z - \gamma_{z-1}) max_{a}Q_{\gamma_{z-1}}(s_{t+1}, a) + \gamma_z max_{a} W_z(s_{t+1}, a)\right]
\end{align}
This Eq. illustrates that the update for $W_z$ incorporates the discrepancy in discount factors $\gamma_z - \gamma_{z-1}$ with respect to the action-value function $Q_{\gamma_{z-1}}(s_{t+1}, a)$ and the bootstrapping from the following step $W_z(s_{t+1}, a)$. This is a Bellman Eq. for $W_{z}$, which combines a decay factor $\gamma_{z}$ and the reward $Q_{\gamma_{z-1}}(s_{t+1}, a)$. Therefore, it can be used to define the predicted TD update for $W_{z}$. In this formulation, $Q_{\gamma_{z-1}}(s_{t+1}, a)$ may be represented as the summation of $W_{z}(s_{t+1}, a_{t+1})$ for $z \leq Z - 1$, suggesting that the Bellman Eq. for $W_{z}$ is dependent upon the values of all delta functions $W_{z}$ for $z \leq Z - 1$.\\
This method considers the delta value function at each time-scale as an autonomous RL problem, with rewards derived from the action-value function of the preceding time-scale. Consequently, with a target discounted action-value function $Q_{\gamma_{z}}(s, a)$, all delta components can be concurrently trained by a TD update, employing the previous values of each estimator for bootstrapping. This method requires the assumption of a series of discount factors, represented as $\gamma_{z}$, encompassing both the minimum and maximum values, $\gamma_{0}$ and $\gamma_{z}$\cite{romoff2019separating}.
\subsection{Multi-Step TD (Q($\Delta$)-Learning)}
\label{mstd-Q}
In normal multi-step TD learning, rather than calculating the temporal difference error based on a single step, we aggregate the rewards over several stages and subsequently bootstrap from the action-value at the final phase. Multiple studies demonstrate that multi-step TD approaches often display superior efficiency relative to single-step TD methods \cite{sutton1998introduction}. In multi-step TD Q($\Delta$)-Learning, the action at each stage is selected greedily according to the highest Q-value. The multi-step temporal difference formula for Q($\Delta$)-Learning is:
\begin{align}
\label{MSTDQ}
W_z(s_t, a_t) = \mathbb{E} \left[ \sum_{j=1}^{k_z-1} (\gamma_z^j - \gamma_{z-1}^j) r_{t+j} + (\gamma_z^{k_z} - \gamma_{z-1}^{k_z}) max_{a}Q_{\gamma_{z-1}}(s_{t+k_z}, a) + \gamma_z^{k_z} W_z(s_{t+k_z}, a_{t+k_z}) \right]
\end{align}
In this context, $r_{t+j}$ denotes the reward acquired at time step $t+j$. The action $a_{t+k_z}$ is selected greedily as $a = \arg\max_{a} Q(s_{t+k_z}, a)$, where $k_z$ corresponds to the number of steps associated with the discount factor $\gamma_z$. This approach extends Q-learning to multi-step TD learning by consolidating rewards over multiple stages and employing bootstrapping from both current and previous time-scale action-value functions. In conventional multi-step TD learning, the TD error is calculated by summing rewards across multiple stages instead of depending on a solitary future step, subsequently bootstrapping from the action-value at the last step. For instance, employing a single discount factor $\gamma$, the multi-step TD update Eq. is expressed as:
\begin{align}
Q_{\gamma}(s_t, a_t) = \mathbb{E}\left[\sum_{i=0}^{k-1} \gamma^i r_{t+i} + \gamma^k \max_{a} Q_{\gamma}(s_{t+k}, a)\right]
\end{align}
where the initial section aggregates rewards across several phases, adjusted by $\gamma$. The second component derives from the largest action-value in the terminal state $s_{t+k}$. We now expand the preceding equation to include numerous discount factors $\gamma_z$ and $\gamma_{z-1}$ by constructing a delta function $W_z(s_t, a_t)$, which denotes the disparity between action-value functions for successive discount factors:
\begin{align}
\label{f-multi-TD-Q}
W_z(s_t, a_t) = Q_{\gamma_z}(s_t, a_t) - Q_{\gamma_{z-1}}(s_t, a_t)\nonumber
\shortintertext{We now execute the multi-step variation of the Bellman Eq. for both $Q_{\gamma_z}(s_t, a_t)$ and $Q_{\gamma_{z-1}}(s_t, a_t)$. The multi-step Bellman Eq. for $Q_{\gamma_z}(s_t, a_t)$ is stated as follows:}
Q_{\gamma_z}(s_t, a_t) = \mathbb{E}\left[\sum_{j=0}^{k_z - 1} \gamma_z^j r_{t+j} + \gamma_z^{k_z} max_{a}Q_{\gamma_z}(s_{t+k_z}, a)\right]\nonumber
\shortintertext{The multi-step Bellman Eq. for $Q_{\gamma_{z-1}}(s_t, a_t)$ is presented as follows:}
Q_{\gamma_{z-1}}(s_t, a_t) = \mathbb{E}\left[\sum_{j=0}^{k_z - 1} \gamma_{z-1}^j r_{t+j} + \gamma_{z-1}^{k_z} max_{a}Q_{\gamma_{z-1}}(s_{t+k_z}, a)\right]\nonumber
\shortintertext{We will now employ the delta estimator formulas to subtract these two Eqs.}
W_z(s_t, a_t) = \left[\sum_{j=0}^{k_z - 1} \gamma_z^j r_{t+j} + \gamma_z^{k_z} max_{a}Q_{\gamma_z}(s_{t+k_z}, a)\right] - \left[\sum_{j=0}^{k_z - 1} \gamma_{z-1}^j r_{t+j} + \gamma_{z-1}^{k_z} max_{a}Q_{\gamma_{z-1}}(s_{t+k_z}, a)\right]\nonumber
\shortintertext{We will now expand and clarify the terms. The immediate reward terms $r_{t+j}$ appear in both Eqs. but are modified by different discount factors, $\gamma_z$ and $\gamma_{z-1}$. This allows us to state the difference in rewards as:}
\sum_{j=0}^{k_z - 1} \gamma_z^j r_{t+j} - \sum_{j=0}^{k_z - 1} \gamma_{z-1}^j r_{t+j} = \sum_{j=0}^{k_z - 1} (\gamma_z^j - \gamma_{z-1}^j) r_{t+j}\nonumber
\shortintertext{We then use the recursive relationship as described for the bootstrapping terms:}
Q_{\gamma_z}(s_{t+k_z}, a) = W_z(s_{t+k_z}, a) + Q_{\gamma_{z-1}}(s_{t+k_z}, a)\nonumber
\shortintertext{Consequently, we obtain:}
\gamma_z^{k_z} max_{a}Q_{\gamma_z}(s_{t+k_z}, a) - \gamma_{z-1}^{k_z} max_{a}Q_{\gamma_{z-1}}(s_{t+k_z}, a) = (\gamma_z^{k_z} - \gamma_{z-1}^{k_z}) max_{a}Q_{\gamma_{z-1}}(s_{t+k_z}, a) + \gamma_z^{k_z} max_{a}W_z(s_{t+k_z}, a)\nonumber
\shortintertext{The final expression is obtained by combining the elements for both the rewards and bootstrapping:}
W_z(s_t, a_t) = \mathbb{E}\left[\sum_{j=0}^{k_z - 1} (\gamma_z^j - \gamma_{z-1}^j) r_{t+j} + (\gamma_z^{k_z} - \gamma_{z-1}^{k_z}) max_{a}Q_{\gamma_{z-1}}(s_{t+k_z}, a) + \gamma_z^{k_z} W_z(s_{t+k_z}, a_{t+k_z})\right]
\end{align}
The objective is to define the update for $W_z$ in a multi-step manner that considers multiple time steps and includes the Q-learning maximization phase. This Eq. expands the traditional multi-step TD update to incorporate multiple discount factors and action-value functions in RL. Consequently, each $W_{z}$ receives a portion of the rewards from the environment up to time-step $k_{z}-1$. Moreover, each $W_z$ employs a unique action-value function in conjunction with the value from the previous time scale. A modified version of this approach, employing k-step bootstrapping as outlined in \cite{sutton1998introduction}, is demonstrated in Algorithm 1. Although Algorithm 1 demonstrates quadratic complexity concerning Z, it may be performed with linear complexity for large Z by maintaining $\hat{Q}$ action-values at each time-scale $\gamma_{z}$.
\begin{algorithm}[H]
   \caption{Multi-Step TD ( Q($\Delta$)-Learning )}
   \label{alg:1}
\begin{algorithmic}
   \State  {\bfseries Inputs:} Choose the discount factors ($\gamma_0, \gamma_1, ..., \gamma_Z$), bootstrapping steps ($k_0, k_1, ..., k_Z$), and learning rates ($\alpha_0, \alpha_1, ..., \alpha_Z$).
   \State Initialize the value of $W_z(s, a) = 0$ for all states, actions, and scales z.
   \For{episode = 0, 1, 2,...}\Comment{Loops for each episode}
   \State Initialize state $s_0$ and determine an initial action $a_0$ based on the $\epsilon$-greedy policy.
   \For{t = 0, 1, 2,...} \Comment{Loops for each time step}
   \State Execute action $a_t$, observe reward $r_t$ and subsequent state $s_{t+1}$.
   \State Choose action $a_{t+1}$ according to the $\epsilon$-greedy policy based on Q(s, a).
   \For{z = 0, 1, ..., Z}
   \If{z = 0}
   \State $G^0 = \sum_{j=0}^{k_0 - 1} \gamma_0^j r_{t+j} + \gamma_0^{k_0} max_{a}W_0(s_{t+k_0}, a)$
   \Else \Comment{By employing Eq.\ref{act-val-fun}, we replace $Q_{\gamma_{z-1}}(s_{t+k_z}, a_{t+k_z})$ by aggregating the W-components up to $W_{z-1}$ in Eq.\ref{f-multi-TD-Q}.}
    \State $G^z = \sum_{j=0}^{k_z - 1} (\gamma_z^j - \gamma_{z-1}^j) r_{t+j} + (\gamma_z^{k_z} - \gamma_{z-1}^{k_z}) \sum_{z=0}^{z-1} max_{a}W_{z-1}(s_{t+k_z}, a) + \gamma_z^{k_z} max_{a}W_z(s_{t+k_z}, a)$
   \EndIf

   \EndFor
   \For {z = 0, 1, 2,..., Z}
   \State $W_z(s_t, a_t) \leftarrow W_z(s_t, a_t) + \alpha_z (G^z - W_z(s_t, a_t))$
   \EndFor
   \EndFor
   \EndFor
\end{algorithmic}
\end{algorithm}

\subsection{Q-Learning TD($\lambda, \Delta$)}
\label{Q-TD-LAMBDA-Delta}
The $\lambda$-return\cite{sutton1984temporal,sutton2018reinforcement} is introduced in Eq.\ref{STD-Return}. It creates a target for TD($\lambda$) updates by combining rewards across a number of steps. The definition of the $\lambda$-return $ G_{t}^{\gamma,\lambda}$ is:
\begin{equation}
\label{STD-Return}
   G_{t}^{\gamma,\lambda}(s_t,a_t) = \hat{Q}_{\gamma}(s_{t}, a_{t})+ \sum_{k=0}^{\infty}(\lambda\gamma)^{k}\delta_{t+k}^{\gamma}
\end{equation}
The TD($\lambda$) operator, defined in Eq.\ref{bellmaneq}, iteratively applies $\lambda$-returns to update the value functions. In Q-Learning, the TD($\lambda$) operator modifies the action-value function by adding the $\lambda$-discounted TD errors, as seen below:
\begin{equation}
\label{bellmaneq}
\mathrm{T}_{\lambda}Q(s_t, a_t) = Q(s_t, a_t) + (I - \lambda \gamma P)^{-1}( \mathrm{T}Q(s_t, a_t) - Q(s_t, a_t))
\end{equation}
Here, P represents the transition matrix for state-action pairs calculated using $max_{a} Q(s_{t+1}, a)$ rather than a policy-based action, while Q is the action-value function. $\mathrm{T}Q(s_t, a_t) = r_{t} + \gamma \max_{a} Q(s_{t+1}, a)$ is the conventional TD(0) update for the action-value function, whereas $(I - \lambda \gamma P)^{-1}$ is the aggregation of multi-step returns, weighted by $\lambda$ and $\gamma$.\\
Likewise, Eq.\ref{SRD-Estimator} describes the $\lambda$-return that is unique to delta estimators in TD($\lambda, \Delta$), shown as $G_{t}^{z,\lambda_{z}}$ for each $W_{z}$:
\begin{equation}
\label{SRD-Estimator}
   G_{t}^{z,\lambda_{z}} := \hat{W}_{z}(s_{t}, a_{t})+ \sum_{k=0}^{\infty}(\lambda_{z}\gamma_{z})^{k}\delta_{t+k}^{z}
\end{equation}
where $\delta_{t}^{0}:=\delta_{t}^{\gamma_{0}}$ and $\delta_{t}^{z} := (\gamma_{z} - \gamma_{z-1} )max_{a}\hat{Q}_{\gamma_{z-1}}(s_{t+1}, a) + \gamma_{z} max_{a}\hat{W}_{z}(s_{t+1},a) - \hat{W}_{z}(s_{t+1},a_{t+1})$ are the TD-errors.
\subsection{Q-Learning TD($\lambda, \Delta$) with Generalized Advantage Estimation (GAE)}
\label{GAE}
Generalized Advantage Estimation (GAE) (\cite{schulman2017proximal}) aims to calculate the advantage function by aggregating multi-step TD errors. To use this strategy with delta estimators $W_{z}(s_{t}, a_{t})$, we compute advantage estimates $A^{\Delta}(s_{t}, a_{t})$ for each time-scale. Each advantage estimate $A^{\Delta}(s_{t}, a_{t})$ uses a multi-step TD error specific to the delta estimator $W_z$, written as follows:
\begin{equation}
\label{sgae}
   A^{\Delta}(s_{t}, a_{t}) = \sum_{k=0}^{T-1}(\lambda_{z}\gamma_{z})^{k}\delta_{t+k}^{\Delta}
\end{equation}
where $\delta_{t+k}^{\Delta} := r_{t} + \gamma_{z}\sum_{z=0}^{Z}max_{a}\hat{W}_{z}(s_{t+1},a) - \sum_{z=0}^{Z}\hat{W}_{z}(s_{t}, a_{t})$.\\
We use $\gamma_{Z}$ as the discount factor, and the total of all W estimators as a substitute for $Q_{\gamma z}$. This aim applies to PPO by performing the policy update from Eq.\ref{cliplossfun} and replacing A with $A^{\Delta}$. In addition, we train each $W_{z}$ using a shortened form of their corresponding $\lambda$-return, as indicated in Eq.\ref{SRD-Estimator}. For further information, see Algorithm \ref{alg:2}.

\begin{algorithm}[H]
   \caption{PPO-TD($\lambda$, Q($\Delta$)-Learning)}
   \label{alg:2}
\begin{algorithmic}
   \State  {\bfseries Inputs:} Choose the discount factors ($\gamma_0, \gamma_1, ..., \gamma_Z$), bootstrapping steps ($k_0, k_1, ..., k_Z$), and learning rates ($\alpha_0, \alpha_1, ..., \alpha_Z$).
   \State Initialize the value of $W_z(s, a) = 0$ for all states, actions, and scales z.
   \State Initialize policy $\omega$ and values $\theta^{z} \forall z$.
   \For{episode = 0, 1, 2,...}\Comment{Loops for each episode}
   \State Initialize state $s_0$ and select an initial action $a_0$ based on a $\epsilon$-greedy policy.
   \For{t = 0, 1, 2,...} \Comment{Loops for each time step}
   \State Execute action $a_t$, observe reward $r_t$, and the subsequent state $s_{t+1}$.
   \State Choose $a_{t+1}$ as the subsequent action using a $\epsilon$-greedy policy.
   \For{z = 0, 1, ..., Z}
   \If{$t \geq T$}
   \State $G^{z, \lambda_z} \leftarrow \hat{W}_z(s_{t-T}, a_{t-T}) + \sum_{k=0}^{T-1} (\lambda_z \gamma_z)^k \delta_{t-T+k}^z \forall z$ \Comment{Calculating the multi-step return $G^{z, \lambda_z}$ and the TD-error $\delta_{t-T+k}^z$ utilizing Eq.\ref{SRD-Estimator}.}
   \EndIf
   \EndFor
   \For {z = 0, 1, 2,..., Z}
   \State $\hat{W}_z(s_{t-T}, a_{t-T}) \leftarrow \hat{W}_z(s_{t-T}, a_{t-T}) + \alpha_z \left(G^{z, \lambda_z} - \hat{W}_z(s_{t-T}, a_{t-T})\right)$
   \State $A^{\Delta} = \sum_{k=0}^{T-1} (\lambda_z \gamma_z)^k \delta_{t-T+k}^{\Delta}$ \Comment{Where $A^{\Delta}$ and $\delta_{t-T+k}^{\Delta}$ are calculated using Eq.\ref{sgae}.}
   \State $\theta^z \leftarrow \theta^z + \alpha_z \left( G^{z, \lambda_z} - \hat{W}_z(s_{t-T}, a_{t-T}) \right) \nabla \hat{W}_z(s_{t-T}, a_{t-T})$ \Comment{Update $\theta^z$ utilizing TD (Eq.\ref{lossfun}) with $G^{z, \lambda_z} \forall z$.}
   \State $\mathcal{L}(\omega) = \mathbb{E}\bigg[min\bigg(\rho(\omega)A^{\Delta}(s,a),clip(\rho(\omega), 1-\epsilon, 1+\epsilon)A^{\Delta}(s,a)\bigg)\bigg]$ \Comment{from Eq.\ref{cliplossfun}}
   \State $\omega \leftarrow \omega + \alpha_\omega \nabla_\omega \mathcal{L}(\omega)$ \Comment{Modify the policy parameters $\omega$ utilizing PPO (Eq.\ref{cliplossfun}) for Q-Learning with $A^{\Delta}$.}
   \EndFor
   \EndFor
   \EndFor
\end{algorithmic}
\end{algorithm}

\section{Analysis}
\label{TA}
We then focus on the trade-off between bias and variance in Q-Learning. A bias-variance trade-off is introduced by the update rule in classical Q-Learning, which mostly depends on bootstrapping the value of the next state. In particular, a larger discount factor ($\gamma$) increases variance because of the accumulation of errors from long-term forecasts, but it also reduces bias by giving future rewards more weight. A smaller discount factor ($\gamma$), on the other hand, minimizes variance by restricting long-term forecasts while focusing more on immediate rewards, leading to larger bias. Action-value components $W_z$ in TD($\Delta$) with higher $\gamma_z$ are linked to larger long-term rewards. But reward sequences are unpredictable, this also results in more variance. We first examine the scenario where the Q-value function matches that of regular Q-Learning, as described in Theorem \ref{thm:thm1}, in order to have a better understanding of when Q($\Delta$)-Learning with multiple discount factors is equivalent to normal Q-Learning ($\hat{Q}_{\gamma}$) under specific conditions. Next, we examine the bias-variance trade-off that results from employing several discount factors and bound the error of the Q($\Delta$)-learning estimator. Together with Kearns and Singh's \cite{kearns2000bias}, these results enable us to use TD($\Delta$) (i.e., Theorem \ref{thm:thm4}), which is provided in terms of bias and variance, to determine an error bound for Q-Learning with various discount factors. To maximize performance, this analysis provides insightful information on how to balance these two quantities \cite{romoff2019separating}.
\subsection{Q-Learning Equivalency Configurations and Enhancements}
\label{SECE}
In certain scenarios, it can be demonstrated that Q-Learning utilizing multiple delta estimators (i.e., employing various discount factors) and its adaptations are equivalent to the conventional Q-Learning estimator $Q(s, a)_{\gamma}$ when reformulated into a value function. Here, we pay particular attention to the Q-value function's linear function approximation, a method frequently employed in Q-Learning for extensive state-action spaces. The primary concept of Q-Learning utilizing multiple delta estimators involves the decomposition of the Q-value function into various components. Each component is associated with a delta estimator defined as $W_z = Q_{\gamma_z}(s, a) - Q_{\gamma_z-1}(s, a)$, which is derived from a sequence of discount factors. The objective is to independently learn these components and subsequently recompose them to capture a wider array of reward structures present in the environment. A linear function approximation is represented in the following form:\\
$\hat{Q}(s, a)_{\gamma} := \langle \theta^{\gamma}, \phi(s, a) \rangle \quad \text{and} \quad \hat{W}_z(s, a) := \langle \theta^z, \phi(s, a) \rangle, \forall z$\\
In this context, $\theta$ and $\theta^{z}$ denote weight vectors within the d-dimensional real space, $\mathbb{R}^d$. The function $\phi: S \times A \rightarrow \mathbb{R}^d$ defines a mapping that associates each state-action pair with a corresponding feature in the specified d-dimensional feature space. Let $Q(s, a)_{\gamma}$ represent the true Q-value function associated with the discount factor $\gamma$, while $\hat{Q}(s, a)_{\gamma}$ denotes the estimated Q-value function utilized by the Q-Learning algorithm, which incorporates multiple delta estimators $W_z(s,a)$. The weight vector $\theta$ for Q-Learning is updated using the TD($\lambda$) learning rule as follows:
\begin{align}
\label{Delta-Gae}
   \theta_{t+1}^{\gamma} = \theta_{t}^{\gamma} + \alpha \bigg(G_{t}^{\gamma, \lambda} - \hat{Q}_{\gamma}(s, a)\bigg)\phi(s_{t},a_{t}),
\end{align}
Here, $G_{t}^{\gamma,\lambda}$ denotes the TD($\lambda$) return as stated in Eq.\ref{STD-Return}. Similarly, the delta estimator method TD($\lambda_{z},\Delta$) is employed to modify each $\hat{W}_{z}$:
\begin{align}
\label{Deltaestimator}
   \theta_{t+1}^{z} = \theta_{t}^{z} + \alpha^z \bigg(G_{t}^{Z, \lambda_{z}} - \hat{W}_{z}(s_{t}, a_{t})\bigg)\phi(s_{t},a_{t}),
\end{align}
where $G_{t}^{Z, \lambda_{z}}$ is defined in the same manner as the TD($\Delta$) return outlined in Eq.\ref{SRD-Estimator}, adapted for the specific action-value function. In these Eqs., $\alpha$ and $\alpha^z$ denote positive learning rates for the standard and delta estimators, respectively. This theorem illustrates the equivalence of two Q-Learning version algorithms in this circumstance. The following theorems are similar to those proposed by Romoff et al. \cite{romoff2019separating}; however, we provide a proof pertaining to the action-value function in the context of Q-Learning, whereas they established the proof for the value function.
\begin{theorem}
\label{thm:thm1}
If $\alpha_{z} = \alpha$, $\lambda_{z}\gamma_{z} = \lambda_{\gamma}$ for all $z$, and if the initial conditions are selected such that $\sum_{z=0}^{Z}\theta_{0}^{\gamma_{z}} = \theta_{0}^{\gamma}$, then the iterates generated by TD($\lambda$) (Eq. \ref{Delta-Gae}) and TD($\lambda, \Delta$) (Eq. \ref{Deltaestimator}) utilizing linear function approximation fulfill the subsequent conditions:
\begin{align}
\label{thm:eq1}
\Sigma_{z=0}^{Z}\theta_{t}^{z} = \theta_{t}^{\gamma}
\end{align}
\end{theorem}
The proof is presented in appendix \ref{proofs}. The initial estimates of the Q-values over different time scales are shown here by $\theta_{0}^{\gamma_{z}}$, while the first estimate for conventional Q-learning is indicated by $\theta_{0}^{\gamma}$.\\
Equivalence is achieved when the learning rates $\alpha_z$ for each delta estimator are identical to the learning rate $\alpha$ employed in conventional Q-Learning, and the product of the learning rates and discount factors satisfies $\lambda_{z}\gamma_{z}=\lambda_{\gamma},\forall z$, thereby ensuring that the updates for the delta estimators correspond with the standard Q-Learning update. As $\lambda$ approaches 1 and $\gamma_{z} < \gamma$, the latter condition indicates that $\lambda_{z} = \lambda_{\gamma}/\gamma_{z}$ may surpass one. It may be deduced that the TD($\lambda_{z}$) may diverge. The following theorem demonstrates that the TD($\lambda$) operator in Eq.\ref{bellmaneq} serves as a contraction mapping for the interval $1 \leq \lambda < \frac{1+\gamma}{2\gamma}$, indicating that $\lambda_{\gamma}<1$\cite{romoff2019separating}.

\begin{theorem}
\label{thm:thm2}
$\forall \lambda \in \left[0, \frac{1 + \gamma}{2\gamma} \right]$, the operator $T_{\lambda}$ for Q-Learning is defined by the Eq. $ T_{\lambda} Q = Q + (I - \lambda \gamma P)^{-1} (T Q - Q),\quad \forall Q \in \mathbb{R}^{|S| \times |A|}$, is a well-defined set. Additionally, the contraction coefficient of $T_{\lambda} Q$ is $\frac{\gamma}{|1 - \lambda \gamma|}$ in relation to the max norm.
\end{theorem}
The proof is presented in appendix \ref{proofs}.\\
Likewise, we may examine the learning of each $W_z$ (delta estimator associated with a distinct discount factor $\gamma_z$) by employing a $k_z$-step TD($\Delta$) rather than the conventional TD($\lambda, \Delta$) update in Q-Learning. The analysis from Theorem \ref{thm:thm1} regarding the value function can be extended to demonstrate that, with linear function approximation, multi-step TD and multi-step TD($\Delta$) are equal if $k_z = k$ for all z. Nonetheless, we see that the equivalence with the unaltered Q-Learning (employing conventional multi-step TD) is the anomaly rather than the standard. Multiple variables contribute to this: (i) Learning Rate and Time-Scales: To establish equivalence between multi-step TD and multi-step TD($\Delta$), it is necessary to maintain a uniform learning rate $\alpha_z = \alpha$ across all time scales. This constitutes a significant limitation, as it seems intuitive that shorter time scales (associated with smaller discount factors $\gamma_z$) can be acquired more rapidly than their longer counterparts. Shorter time scales typically necessitate fewer updates to accurately reflect the dynamics of immediate rewards, whereas longer time scales demand more updates to account for the cumulative impacts of delayed rewards. (ii) Adaptive Optimizers: In practice, adaptive optimizers (e.g., Adam, RMSprop) are frequently employed in the context of nonlinear approximation (e.g., in deep Q-learning)\cite{henderson2018did, schulman2017proximal}. These optimizers modify the learning rate according to the characteristics of each delta estimator and its corresponding target. The effective learning rate may vary based on the characteristics of each delta estimator and its corresponding goal. Consequently, the optimizer can autonomously adjust the learning rates to vary over shorter and longer time scales. (iii) Decomposition of the Q-value Function: The decomposition of the Q-value function into delta components provides advantageous qualities that the non-delta estimator (ordinary Q-Learning) does not offer. Each delta component $W_z$ does not necessarily have to utilize the same k-step return or TD($\lambda$)-return as the non-delta estimator or the superior components $W_z$. If $k_z < k_{z+1}, \forall z \quad (or \quad \gamma_{z}\lambda_{z} < \gamma_{z+1}\lambda_{z+1}, \forall z)$, variance reduction is feasible, but perhaps accompanied by some bias introduction. (iv) Variance Reduction and Bias Introduction: Employing shorter time scales (diminished $\gamma_z$ and $\lambda_z$ values) for the lower components $W_z$ facilitates a reduction in the variance of the Q-value estimates. This is particularly advantageous in contexts characterized by high stochasticity or sparse rewards, because short-term forecasts may be rendered with greater confidence, hence diminishing overall variance. Nevertheless, shorter time scales may introduce some bias, as they may not completely capture the long-term reward structure. This trade-off necessitates meticulous management through the adjustment of discount factors and step sizes to achieve an ideal equilibrium between bias and variation.
\subsection{Evaluation for Minimizing $k_{z}$ Values in Q-Learning via Phased Updates}
\label{arv}
To see how Q-Learning with multiple discount factors (Q ($\Delta$)-Learning) differs from the single estimator situation (standard Q-Learning), consider the tabular phased form of k-step Q-Learning, which is comparable to the k-step TD approach investigated by Kearns \& Singh \cite{kearns2000bias}. In the single estimator scenario, we commonly use a single discount factor $\gamma$ and update the Q-values based on the TD error and discounted return over a given number of steps. This strategy can result in significant variance in environments with delayed rewards or limited feedback.

In this context, we generate n trajectories by starting from each state $s \in \mathcal{S}$ and performing an action $a \in \mathcal{A}$. Each trajectory is denoted as:
$\bigg\{ S_{0}^{(j)} = s, a_{0}, \gamma_{0}, ..., S_{k}^{(j)}, a_{k}^{(j)}, \gamma_{k}^{(j)}, S_{k+1}^{(j)}, ...\bigg\}_{1\leq j \leq n}$
The trajectories are formed by following a policy $\pi$, which may be an exploratory or greedy policy depending on the current Q-value estimates. In phased multi-step Q-Learning with Multiple Delta Estimators, the updates for each component $W_z$, which integrate rewards up to $k_{z-1}$, are executed at each phase t be defined as:
\begin{align}
\hat{Q}_{\gamma_z,t}(s_t, a_t) = \frac{1}{n} \sum_{j=1}^n \left[ \sum_{i=0}^{k_z-1} \gamma_z^i r_i(j) + \gamma_z^{k_z} \max_{a} \hat{Q}_{\gamma_z,t-1}(s(k_j), a) \right]
\end{align}
Theorem \ref{thm:thm3} presented below is derived from the work of Kearns \& Singh\cite{kearns2000bias}, who focus on the state value function. In contrast, we provide an upper bound on the error for Q($\Delta$)-Learning utilizing multiple delta estimators defined by $\Delta \hat{Q}_{\gamma,t} := \max_{s} \left| \max_{a} \hat{Q}_{\gamma,t}(s, a) - \max_{a} Q_{\gamma}(s, a) \right|$

\begin{theorem}
\label{thm:thm3}
Motivated by Kearns and Singh \cite{kearns2000bias}, for any $0<\delta<1$, define $\epsilon = \sqrt{\frac{2\log(2k/\delta)}{n}}$, which holds with a probability of $1-\delta$,
\begin{align}
\label{thm3:Eq}
\Delta_{t}^{\hat{Q}_{\gamma}} \leq \epsilon \bigg(\underbrace{\frac{1-\gamma^{k}}{1-\gamma} \bigg)}_{\text{variance term}}+ \underbrace{\gamma^{k} \Delta_{t-1}^{\hat{Q}_{\gamma}}}_{\text{bias term}}
\end{align}
\end{theorem}
The proof is presented in appendix \ref{proofs}.\\
Where $\Delta_t^{\hat{Q}_{\gamma}}$ denotes the discrepancy between the estimated Q-value at time step t and the true Q-value. In Eq. \ref{thm3:Eq}, a variance term ($\frac{1-\gamma^{k}}{1-\gamma}$) emerges as a result of sampling error from rewards obtained along trajectories. Specifically, $\epsilon$ constrains the divergence of the empirical average of rewards from the actual expected reward. The second term ($\gamma^{k} \Delta_{t-1}^{\hat{Q}_{\gamma}}$) in Eq. \ref{thm3:Eq} represents a bias that originates from bootstrapping the estimations of the preceding phase. Likewise, we examine a phased Q-Learning variant of multi-step TD($\Delta$). For every phase t, we update each W accordingly:
\begin{align}
\hat{W}_{z,t}(s,a) = \frac{1}{n}\sum_{j=1}^{n}\bigg(\sum_{i=1}^{k-1}(\gamma_{z}^{i} - \gamma_{z-1}^{i})r_{i}^{(j)} + (\gamma_{z}^{k_{z}} - \gamma_{z-1}^{k_{z}})max_{a}Q_{\gamma_{z-1}}(s_{t+k}^{(j)}, a) + \gamma_{z}^{k_{z}} \hat{W}_{z}(s_{t+k}^{(j)}, a_{t+k}^{(j)})
\end{align}
We now determine an upper limit on the phased error. $TD(\Delta)$ is defined as the cumulative error associated with each component $W$, expressed as $\sum_{z=0}^{Z}\Delta_{t}^{z}$, where $\Delta_{t}^{z}= \max_{s, a}\bigg\{\bigg | \hat{W}_{z}(s,a) - W_{z}(s,a) \bigg |\bigg\}$. Principal distinctions Between SARSA and Q-learning regarding the error bound: In SARSA, the error is determined by the actual action executed by the agent, indicating that we update the Q-values according to the state-action pairs the agent encounters during exploration. In Q-learning, the error is determined by the greedy action, which is the action that maximizes the Q-value for the subsequent state, under the assumption that the agent consistently acts optimally.
\begin{theorem}
\label{thm:thm4}
Let it be assumed that $\gamma_{0} \leq \gamma_{1} \leq \gamma_{2} \leq \ldots \leq \gamma_{z} = \gamma$. and $K_{0} \leq K_{1} \leq ... \leq K_{z}= K$, for any $0 < \delta < 1$, define $\epsilon = \sqrt{\frac{2\log(2k/\delta)}{n}}$, with a probability of $1 - \delta$,
\begin{align}
\label{thm4:eq1}
\sum_{z=0}^{Z} \Delta_{t}^{z} \leq \epsilon \bigg( \frac{1-\gamma^{k}}{1 - \gamma} \bigg)+ \epsilon \bigg(\underbrace{\sum_{z=0}^{Z-1} \frac{\gamma_{z}^{k_{z+1}}-\gamma^{k_{z}}_{z}}{1-\gamma_{z}} \bigg)}_{\text{variance reduction}}+ \underbrace{\sum_{z=0}^{Z-1}\bigg( \gamma^{k_{z}}_{z} - \gamma^{k_{z+1}}_{z} \bigg)\sum_{u=0}^{z}\Delta_{t-1}^{u}}_{\text{bias introduction}} + \gamma^{k}\sum_{z=0}^{Z}\Delta_{t-1}^{z}
\end{align}
\end{theorem}
The proof is presented in appendix \ref{proofs}.\\
The error bound for Q-learning with phased TD($\Delta$) follows a similar structure to the error bound for value functions, but now it's applied to action-value functions (Q-values). We will now juxtapose the error bounds for Q-learning utilizing phased TD($\lambda$) as presented in theorem \ref{thm:thm3} and phased TD($\Delta$) as defined in theorem \ref{thm:thm4}. In Q-learning employing phased TD($\Delta$), the error bound for each delta estimator encompasses components related to variance reduction and bias introduction.
The error bound for phased TD($\Delta$) includes a term facilitating variance reduction, governed by the differences in the discount factors $\gamma_z^{k_{z+1}} - \gamma_z^{k_z}$, which is equivalent to $\epsilon \left( \sum_{z=0}^{Z-1} \frac{\gamma_z^{k_{z+1}} - \gamma_z^{k_z}}{1 - \gamma_z} \right) \leq 0$. In phased TD($\Delta$) for Q-learning, diminishing the step sizes $k_z$ to attain variance reduction may result in the introduction of bias quantified as $\sum_{z=0}^{Z-1} \left( \gamma_z^{k_z} - \gamma_z^{k_{z+1}} \right) \sum_{u=0}^{z} \Delta_{t-1}^u \geq 0$. This results from the cumulative bias of the shorter-horizon estimations, which are affected by prior delta estimators. When all $k_{z}$ values are identical, both algorithms produce the same upper bound.

The fundamental element of effective learning in Q-learning with phased TD($\Delta$) is attaining a balance between variance reduction and bias introduction. Overly diminishing $k_z$ to control variance may induce bias, while excessively increasing $k_z$ may result in substantial variance. Phased TD($\Delta$) provides a technique to reduce variance through the average of many delta estimators while ensuring a negligible rise in bias.
Kearns and Singh\cite{kearns2002near} have established in the literature that the expected discounted return over $\mathrm{T}$ steps approximates the infinite-horizon discounted expected return after $T \approx \frac{1}{1-\gamma}$ steps. This outcome indicates that $k_z$ can be diminished for any $\gamma_z$ such that: $k_z \approx \frac{1}{1 - \gamma_z}$. This modification enables adherence to a principle whereby the step sizes $k_z$ correlate with the discount factors $\gamma_z$, facilitating an optimal bias-variance trade-off. Specifically, when (i) $k_z \approx \frac{1}{1 - \gamma_z}$, the influence of longer horizons will be suitably balanced, therefore diminishing both bias and variance. (ii) For substantial $\mathrm{T}$ (the number of updates), this selection of $k_z$ guarantees an optimal balance between variance reduction (via averaging several delta estimators) and minimal bias introduction (by preventing shorter-horizon estimates from prevailing).
\section{Experiments}
\label{experiment}
We validate the performance of Q($\Delta$)-Learning on standard RL benchmarks, including Atari games and the Ring MDP. Our results show that Q($\Delta$)-Learning outperforms traditional Q-Learning in terms of both learning speed and final performance, especially in environments with long-term dependencies.

\section{Conclusion}
\label{conclusion}
Q($\Delta$)-Learning provides a novel way to extend Q-Learning by incorporating delta components across different time scales. This method allows for better control of bias-variance trade-offs and improves performance in long-term reward settings.%

\bibliographystyle{nature} 
\bibliography{main-Q.bib}

\section*{Acknowledgments}

\section*{Author contributions statement}

\section*{Data availability}
The data sets produced in this work can be obtained from the corresponding author upon an appropriate request.
\section*{Competing interests}
The authors disclose no conflicting interests.

\section*{Appendix}
\appendix

\section*{PROOFS}
\label{proofs}
\begin{proof}{Theorem \ref{thm:thm1}}
\begin{align}
\shortintertext{Induction is employed in the proof. First, it is important to prove that the base case is valid at t = 0. This statement is clearly valid based on our initial premise, especially regarding zero initiation. At a certain time-step t, we assert the truth of the proposition, specifically, $\sum_{z=0}^{Z}\theta_{t}^{z} = \theta_{t}^{\gamma}$. We shall now illustrate that it also applies at the next time step, t+1.}\nonumber
\sum_{z=0}^{Z}\theta_{t+1}^{z} &= \sum_{z=0}^{Z}\bigg( \theta_{t}^{z} + \alpha_{z} \bigg(G_{t}^{Z, \lambda_{z}} - \hat{W}_{z}(s_{t}, a_{t})\bigg)\phi(s_{t},a_{t})\bigg)\\ \nonumber
   W_{z}(s, a) &= Q_{\gamma_{z}}(s, a) \mathit{-} Q_{\gamma_{z -1}}(s, a) \text{ and using Eq.\ref{SRD-Estimator}}\\ \nonumber
               &= \theta_{t}^{\gamma} + \sum_{z=0}^{Z}\alpha_{z} \bigg(\sum_{k=t}^{\infty}(\lambda_{z}\gamma_{z})^{k-t}\delta_{k}^{z}\bigg)\phi(s_{t},a_{t})\text { , Induction-based assumption.}\\ \nonumber
               &= \theta_{t}^{\gamma} + \alpha \sum_{k=t}^{\infty}(\lambda \gamma)^{k-t} \underbrace{\bigg( \sum_{z=0}^{Z}\delta_{k}^{z}\bigg)}_{*}\phi(s_{t},a_{t})\text{ , Because of $\alpha_{z} = \alpha$, $\lambda_{z}\gamma_{z} = \lambda_{\gamma}$, $\forall z$}\\
\shortintertext{We must prove that the term (*) = $\sum_{z=0}^{Z}\delta_{k}^{z}$ is equal to the traditional TD error $\delta_{k}^{\gamma}$ in order to demonstrate that $\sum_{z=0}^{Z}\theta_{t+1}^{z} = \theta_{t+1}^{\gamma}$.}\nonumber
\sum_{z=0}^{Z}\delta_{k}^{z} &= r_{k} +\gamma_{0}max_{a}\hat{Q}_{\gamma_{0}}(s_{k+1},a) - \hat{Q}_{\gamma_{0}}(s_{k},a_{k}) + \sum_{z=1}^{Z}\bigg( (\gamma_{z} - \gamma_{z-1})\sum_{u=0}^{z-1}\langle\theta_{t}^{u},\phi(s_{k+1},a_{k+1})\rangle + \gamma_{z}\langle\theta_{t}^{z},\phi(s_{k+1},a_{k+1})\rangle - \langle\theta_{t}^{z},\phi(s_{k},a_{k})\rangle \bigg)\\ \nonumber
&= r_{k} +\gamma_{0}max_{a}\hat{Q}_{\gamma_{0}}(s_{k+1},a) + \bigg\langle \sum_{z=1}^{Z}\bigg(\gamma_{z} \sum_{u=0}^{z}\theta_{t}^{u} - \gamma_{z-1}\sum_{u=0}^{z-1}\theta_{t}^{u}\bigg),\phi(s_{k+1},a_{k+1})\bigg\rangle - \bigg\langle \sum_{z=0}^{Z}\theta_{t}^{z},\phi(s_{k},a_{k})  \bigg\rangle \\ \nonumber
&= r_{k} +\gamma_{0}max_{a}\hat{Q}_{\gamma_{0}}(s_{k+1},a)+ \gamma_{z} \bigg\langle \sum_{z=0}^{Z}\theta_{t}^{z},\phi(s_{k+1},a_{k+1})\bigg\rangle - \gamma_{0}\bigg\langle\theta_{t}^{0},\phi(s_{k+1},a_{k+1})\bigg\rangle - Q_{\gamma}(s_{k},a_{k})\\ \nonumber
&= r_{k} +\gamma_{0}max_{a}\hat{Q}_{\gamma_{0}}(s_{k+1},a) + \gamma max_{a}\hat{Q}_{\gamma}(s_{k+1},a) - \gamma_{0}max_{a}\hat{Q}_{\gamma_{0}}(s_{k+1},a) - \hat{Q}_{\gamma}(s_{k},a_{k})\\ \nonumber
&= r_{k} + \gamma max_{a}\hat{Q}_{\gamma}(s_{k+1},a) - \hat{Q}_{\gamma}(s_{k},a_{k})\\
&= \delta_{k}^{\gamma}
\end{align}
\end{proof}

\begin{proof}{Theorem \ref{thm:thm2}}
\begin{align}
\shortintertext{The demonstration of the contraction property of $\mathrm{T}\lambda$ can be modified for Q-learning via a linear function approximation, analogous to the approach employed for the value function\cite{romoff2019separating}. The primary distinction resides in the characterization of the Q-Learning update rule and the Bellman operator. The Q-Learning update rule for the Q-function is expressed as $Q(s_t, a_t) \leftarrow Q(s_t, a_t) + \alpha \left( r_{t+1} + \gamma \max_{a} Q(s_{t+1}, a) - Q(s_t, a_t) \right)$, with the Bellman operator for Q-Learning denoted as $\mathrm{T} Q(s, a) = r(s, a) + \gamma \sum_{s'} P(s'|s, a) \max_{a} Q(s', a)$. The following is the definition of the Bellman operator:}\nonumber
\mathrm{T} = r + \gamma P \\ \nonumber
\shortintertext{The symbols $\gamma$, r, and P stand for the discount factor, the expected reward function, and the transition probability operator that the policy $\pi$ induces, respectively.The definition of the TD($\lambda$) operator is a geometric weighted sum of $\mathrm{T}$, which may be written as follows:}
\mathrm{T} = (1-\lambda)\sum_{k=0}^{\infty} \lambda^{k}(\mathrm{T})^{k+1}\label{SARSA-Bellman-Operator}\\
\shortintertext{The aforementioned total must be finite, which implies that $\lambda \in [0, 1]$ is required. For each function W, the equivalent definition of $\mathrm{T}_{\lambda}$ is as follows:}
\mathrm{T}_{\lambda}W = W + (I - \lambda \gamma P)^{-1}(\mathrm{T}W - W)\label{SARSA-Sep-Action-Value}\\
\shortintertext{In this form, W represents the separated action-value function, and the update is expressed in terms of the operator $\mathrm{T}_{\lambda}$, which applies time-scale separation. If $0\leq \lambda\gamma<1$, then the formula is regarded as well-defined. Eq. $I - \lambda\gamma P$ is invertible since the spectral norm of the operator $\lambda\gamma P$ is smaller than 1. However, the equivalence between Eqs. \ref{SARSA-Bellman-Operator} and \ref{SARSA-Sep-Action-Value} is lost when $\lambda$ is greater than one. We actually use the TD error for training, which, as we expected, matches the specification of $\mathrm{T}_{\lambda}$ given in Eq. \ref{SARSA-Sep-Action-Value}, thus this is not an issue. The contraction property of the operator $\mathrm{T}_{\lambda}$ will now be examined. The first step is to rewrite the Eq. known as \ref{SARSA-Sep-Action-Value} as follows:}
\mathrm{T}_{\lambda} &= (I - \lambda \gamma P)^{-1}(\mathrm{T}W - \lambda \gamma P W)\\ \nonumber
\shortintertext{Two action-value functions, $W_{1}$ and $W_{2}$, are examined in order to illustrate the contraction property. The updates of $\mathrm{T}_{\lambda}W_1$ and $\mathrm{T}_{\lambda}W_2$ differ in the following ways:} \nonumber
\mathrm{T}_{\lambda}W_{1} - \mathrm{T}_{\lambda}W_{2}& = (I - \lambda \gamma P)^{-1}\bigg(\mathrm{T}W_{1} - \mathrm{T}W_{2} - \lambda \gamma P (W_{1} - W_{2})\bigg)\\ \nonumber
                                                       &= (I - \lambda \gamma P)^{-1}\bigg(\gamma P (W_{1} - W_{2}) - \lambda \gamma P (W_{1} - W_{2})\bigg)\\ \nonumber
                                                       &= (I - \lambda \gamma P)^{-1}\bigg(\gamma (1-\lambda)P (W_{1} - W_{2})\bigg)\\ \nonumber
\shortintertext{We arrive at the following conclusion:} \nonumber
\|\mathrm{T}_{\lambda}W_{1} - \mathrm{T}_{\lambda}W_{2}\| \leq \frac{\gamma|1 - \lambda|}{1 - \lambda\gamma}\|W_{1} - W_{2}\| \textit{ where P = 1}\\ \nonumber
\shortintertext{The fact that $0\leq \lambda\leq 0$ is a contraction is known. The following requirement must be met for $\lambda > 1$:} \nonumber
\frac{\gamma(\lambda - 1)}{1 - \lambda\gamma} < 1 \Rightarrow \gamma\lambda - \gamma < 1 - \lambda\gamma\\
\Rightarrow 2 \lambda\gamma < 1 + \gamma \Rightarrow \lambda < \frac{1 + \gamma}{2\gamma}\\
\shortintertext{As a result, $\mathrm{T}_{\lambda}$ is a contraction if $0 \leq \lambda < \frac{1 + \gamma}{2\gamma}$. This clearly shows that $\gamma\lambda < 1$ for $\gamma < 1$.} \nonumber
\end{align}
\end{proof}

\begin{proof}{Theorem \ref{thm:thm3}}
\begin{align}\label{prf:thm3}\nonumber
\shortintertext{The Q-Learning update rule for the Q-function is stated as $Q(s_t, a_t) \leftarrow Q(s_t, a_t) + \alpha \left( r_{t+1} + \gamma \max_{a} Q(s_{t+1}, a) - Q(s_t, a_t) \right)$.}
\shortintertext{In Q-Learning, we evaluate the $\mathit{Q-value}$ function, Q(s, a), by utilizing samples of state-action pairs together with their associated rewards. Let $Q_{t}(s,a)$ represent the estimate of the $\mathit{Q-value}$ at time t. For $\mathit{n}$ samples, Hoeffding's inequality assures that for a variable restricted within $[-1, +1]$, the chance of the sample mean deviating from the true mean by more than $\epsilon$ is expressed as:}\nonumber
Q^{\pi}(s, a) = \mathbf{E}\bigg[\gamma_{0} + \gamma r_{1} + ... + \gamma^{k-1}r_{k-1} + \gamma^{k}Q^{\pi}(s_{k}, a_{k}) \bigg]\nonumber
\shortintertext{The expectations pertain to a stochastic trajectory under policy $\pi$; therefore, $\mathbf{E}[r_{\iota}](\iota \leq k - 1)$ signifies the predicted value of the $\iota$th reward obtained, while $\mathbf{E}[Q^{\pi}(s_{k}, a_{k})]$ represents the estimated value of the true value function at the $k$th state-action encountered. The phased TD(k) update aggregates the terms $\gamma^{\iota}\bigg(\frac{1}{n}\bigg)\sum_{i=1}^{n}r_{\iota}^{i}$, whose expectations precisely correspond to $\gamma^{\iota}\mathbf{E}[r_{\iota}]$ as referenced in \cite{kearns2000bias}.}\nonumber
\mathbb{P}\bigg(\bigg|\frac{1}{n}\sum_{j=1}^{n} r_{i}^{(j) } \mathit{-} \mathbb{E}[r_{i}] \bigg|\geq\epsilon\bigg) \leq 2e^{\bigg(\mathit{-} \frac{2n^{^{2}}\epsilon^{2}}{\sum_{j=1}^{n}(b \mathit{-} a)^{2}} \bigg)}\textit{where a=-1 and b=+1}\nonumber\\
\mathbb{P}\bigg(\bigg|\frac{1}{n}\sum_{j=1}^{n} r_{i}^{(j) } \mathit{-} \mathbb{E}[r_{i}] \bigg|\geq\epsilon\bigg) \leq 2e^{\bigg(\mathit{-} \frac{2n^{^{2}}\epsilon^{2}}{n2^{2}} \bigg)}\\
\shortintertext{Assuming n and the likelihood of surpassing a $\epsilon$ value is constrained to a maximum of $\delta$, we may determine the resultant value of $\epsilon$:}\nonumber
2e^{\bigg(\mathit{-} \frac{2n^{^{2}}\epsilon^{2}}{n2^{2}} \bigg)} = \delta \nonumber\\
e^{\bigg(\mathit{-} \frac{2n^{^{2}}\epsilon^{2}}{n2^{2}} \bigg)} = \frac{\delta}{2}\nonumber\\
\shortintertext{Applying the Natural Logarithm}\nonumber
ln\bigg(e^{\bigg(\mathit{-} \frac{2n^{^{2}}\epsilon^{2}}{n2^{2}} \bigg)}\bigg) = ln\bigg(\frac{\delta}{2}\bigg)\nonumber\\
\mathit{-} \frac{2n^{^{2}}\epsilon^{2}}{n2^{2}} = ln\frac{\delta}{2}\nonumber\\
\mathit{-} \frac{n\epsilon^{2}}{2} = ln\frac{\delta}{2}\nonumber\\
\mathit{-} \frac{n\epsilon^{2}}{2} = ln\frac{\delta}{2}\nonumber\\
\frac{n\epsilon^{2}}{2} = \mathit{-}ln\frac{\delta}{2}\nonumber\\
\frac{n\epsilon^{2}}{2} = \mathit{-}\bigg(ln(\delta) - ln(2)\bigg)\nonumber\\
\frac{n\epsilon^{2}}{2} = \mathit{-}\bigg(ln(\delta) + ln(2)\bigg)\nonumber\\
\frac{n\epsilon^{2}}{2} = \bigg(ln(2) \mathit{-} ln(\delta)\bigg)\nonumber\\
\frac{n\epsilon^{2}}{2} = ln\frac{2}{\delta}\nonumber\\
n\epsilon^{2} = 2 log\frac{2}{\delta}\nonumber\\
\epsilon^{2} = \frac{2 log\frac{2}{\delta}}{n}\nonumber\\
\epsilon = \sqrt{\frac{2 log\frac{2}{\delta}}{n}}\\
\shortintertext{According to Hoeffding's inequality, for n samples, the probability is at least $1 - \delta$,}\nonumber\\
\mathbb{P}\bigg(\bigg|\frac{1}{n}\sum_{j=1}^{n} r_{i}^{(j) } \mathit{-} \mathbb{E}[r_{i}] \bigg|\bigg) \leq \epsilon = \sqrt{\frac{2 log\frac{2}{\delta}}{n}}\\
\shortintertext{Given that we are addressing k distinct state-action pairings, we employ a union bound. To guarantee the probability applies to all k hypotheses, modify $\delta$ to $\delta/k$:}\nonumber
\mathbb{P}\bigg(\bigg|\frac{1}{n}\sum_{j=1}^{n} r_{i}^{(j) } \mathit{-} \mathbb{E}[r_{i}] \bigg|\bigg) \leq \epsilon = \sqrt{\frac{2 log\frac{2k}{\delta}}{n}}\nonumber\\
\shortintertext{We may now presume that all $\mathbb{E}[r_{i}]$ terms are estimated with a minimum precision of $\epsilon$. By reinserting this into the definition of the k-step TD update, we obtain}\nonumber
\hat{Q}_{t+1}(s, a) \mathit{-} Q(s, a) = \frac{1}{n}\sum_{j=1}^{n}\bigg(r_{0} + \gamma r_{1} + ... + \gamma^{k-1}r_{k-1} + \gamma^{k}Q_{t}(s_{k}, a_{k}) \bigg) \mathit{-} Q(s, a)\nonumber\\
= \sum_{i=0}^{k-1}\gamma^{i} \bigg(\frac{1}{n}\sum_{i=1}^{n}r_{i}^{(j)} -\mathbb{E}[r_{i}] \bigg) + \gamma^{k} \bigg(\frac{1}{n}\sum_{i=1}^{n} Q_{t}(s_{k}^{i}, a_{k}^{i}) -\mathbb{E}[Q(s_{k},a_{k})] \bigg)\\
\shortintertext{In the second line, we reformulated the value as a summation of k rewards. We have now established an upper bound for the difference from $E[r_{i}]$ by $\epsilon$ to obtain}\nonumber
\hat{Q}_{t+1}(s, a) \mathit{-} Q(s, a) \leq \sum_{i=0}^{k-1}\gamma^{l} \epsilon + \gamma^{k} \bigg(\frac{1}{n}\sum_{i=1}^{n} Q_{t}(s_{k}^{i}, a_{k}^{i}) -\mathbb{E}[Q(s_{k},a_{k})] \bigg)\nonumber\\
\shortintertext{The variance term originates from the disparity between the empirical average of rewards and the genuine expected reward.}\nonumber
\epsilon \bigg( \frac{1-\gamma^{k}}{1 - \gamma} \bigg)\nonumber\\
\leq \epsilon \bigg( \frac{1-\gamma^{k}}{1 - \gamma} \bigg) + \gamma^{k} \bigg(\frac{1}{n}\sum_{i=1}^{n} Q_{t}(s_{k}^{i}, a_{k}^{i}) -\mathbb{E}[Q(s_{k},a_{k})] \bigg)\\
\shortintertext{The bias term is disseminated via bootstrapping:}\nonumber
\gamma^{k}\Delta^{\hat{Q}_{\gamma}}_{t-1}\nonumber\\
\shortintertext{The second term is constrained by $\Delta^{\hat{Q}_{\gamma}}_{t-1}$ as per the assumption. Thus, by amalgamating these terms, we derive the comprehensive bound for Q-Learning:}
\Delta^{\hat{Q}_{\gamma}}_{t} \leq \epsilon \bigg( \frac{1-\gamma^{k}}{1 - \gamma} \bigg) + \gamma^{k}\Delta^{\hat{Q}_{\gamma}}_{t-1}
\end{align}
\end{proof}

\begin{proof}{Theorem \ref{thm:thm4}}
\begin{align}\label{prf:thm4}\nonumber
\shortintertext{Phased TD($\Delta$) update criteria for $z \geq 1$:}
\hat{W}_{z,t}(s,a) = \frac{1}{n}\sum_{j-1}^{n}\bigg(\sum_{i=1}^{k_{z}-1} (\gamma_{z}^{i} - \gamma_{z-1}^{i})r_{i}^{(j)} + (\gamma_{z}^{k_{z}} - \gamma_{z-1}^{k_{z}})max_{a^j}\hat{Q}_{\gamma_{z-1}}(s_{k}^{(j)}, a^{(j)}) + \gamma_{z}^{k_{z}} \hat{W}_{z}(s_{k}^{(j)}, a_{k}^{(j)})\bigg)
\shortintertext{It is established that, in accordance with the multi-step update rule \ref{MSTDQ}, for $z \geq 1$:}
W_{z}(s_{t}, a_{t}) = \mathbb{E}\bigg[\sum_{i=1}^{k_{z}-1} \bigg(\gamma_{z}^{i} - \gamma_{z-1}^{i} \bigg)r_{i} + \bigg(\gamma_{z}^{k_{z}} - \gamma_{z-1}^{k_{z}} \bigg) max_{a}Q_{\gamma_{z-1}}(s_{k}, a) + \gamma_{z}^{k_{z}} W_{z}(s_{k},a_{k}) \bigg]\\
\shortintertext{Subsequently, the difference between the two formulas yields for $z \geq 1$:}\nonumber\\
\hat{W}_{z,t}(s,a) - W_{z}(s_{t}, a_{t}) = \sum_{i=1}^{k_{z}-1} \bigg(\gamma_{z}^{i} - \gamma_{z-1}^{i} \bigg)\bigg( \frac{1}{n}\sum_{j-1}^{n}r_{i}^{(j)} - \mathbb{E}[r_{i}]\bigg) + \bigg(\gamma_{z}^{k_{z}} - \gamma_{z-1}^{k_{z}} \bigg) \bigg(\sum_{u=0}^{z-1} \hat{W}_{u}(s_{k}^{(j)},a_{k}^{(j)}) - \mathbb{E}[W_{z}(s_{k},a_{k})]  \bigg)\nonumber\\
+ \gamma_{z}^{k_{z}} \bigg(W_{z}(s_{k}^{(j)},a_{k}^{(j)}) - \mathbb{E}[W_{z}(s_{k},a_{k})] \bigg)
\shortintertext{Assuming $k_{0}\leq k_{1}\leq... k_{z}=k$, the W estimates share no more than $k_{z}=k$ reward terms $\frac{1}{n}\sum_{j-1}^{n}r_{i}^{(j)}$.  By employing the Hoeffding inequality and union bound, we ascertain that with a probability of $1 - \delta$, each k empirical average reward $\frac{1}{n}\sum_{j-1}^{n}r_{i}^{(j)}$ diverges from the genuine expected reward $\mathbb{E}[r_{i}]$ by up to $\epsilon = \sqrt{\frac{2log(2k/\delta)}{n}}$. Thus, with a probability of $1 - \delta$, for every $z \geq 1$, we get:}\nonumber\\
\Delta_{t}^{z} \leq \epsilon \sum_{i=1}^{k_{z}-1} \bigg(\gamma_{z}^{i} - \gamma_{z-1}^{i} \bigg) + \bigg(\gamma_{z}^{k_{z}} - \gamma_{z-1}^{k_{z}} \bigg) \sum_{u=0}^{z-1} \Delta_{t-1}^{u} + \gamma_{z}^{k_{z}} \Delta_{t-1}^{z}\nonumber\\
= \epsilon \bigg(\frac{1-\gamma_{z}^{k_{z}}}{1 - \gamma_{z}} - \frac{1 - \gamma_{z-1}^{k_{z}}}{1 - \gamma_{z-1}} \bigg) + \bigg(\gamma_{z}^{k_{z}} - \gamma_{z-1}^{k_{z}} \bigg) \sum_{u=0}^{z-1} \Delta_{t-1}^{u} + \gamma_{z}^{k_{z}} \Delta_{t-1}^{z}
\shortintertext{ and $\Delta_{t}^{0} \leq \epsilon \frac{1 - \gamma_{0}^{k_{0}}}{1 - \gamma_{0}} + \gamma_{0}^{k_{0}}\Delta_{t-1}^{0}$}
\shortintertext{The addition of the two preceding inequalities yields:}\nonumber
\sum_{z=0}^{Z} \Delta_{t}^{z} \leq  \epsilon \frac{1 - \gamma_{0}^{k_{0}}}{1 - \gamma_{0}} + \epsilon \sum_{z=1}^{Z} \bigg(\frac{1-\gamma_{z}^{k_{z}}}{1 - \gamma_{z}} - \frac{1 - \gamma_{z-1}^{k_{z}}}{1 - \gamma_{z-1}} \bigg) + \sum_{z=1}^{Z} \bigg(\gamma_{z}^{k_{z}} - \gamma_{z-1}^{k_{z}} \bigg) \sum_{u=0}^{z-1} \Delta_{t-1}^{u} + \gamma_{z}^{k_{z}} \Delta_{t-1}^{z}\\
= \underbrace{\epsilon \frac{1-\gamma_{Z}^{k_{Z}}}{1 - \gamma_{Z}} + \epsilon \sum_{z=0}^{Z-1}  \frac{\gamma_{z}^{k_{z+1}}-\gamma_{z}^{k_{z}}}{1 - \gamma_{z}}}_{\text{(*)variance term}}   + \underbrace{\sum_{z=1}^{Z} \bigg(\gamma_{z}^{k_{z}} - \gamma_{z-1}^{k_{z}} \bigg) \sum_{u=0}^{z-1} \Delta_{t-1}^{u} + \gamma_{z}^{k_{z}} \Delta_{t-1}^{z}}_{\text{(**)bias term}}\\
\shortintertext{Let us now concentrate more especially on the bias term (**)}\nonumber\\
\sum_{z=1}^{Z} \bigg(\gamma_{z}^{k_{z}} - \gamma_{z-1}^{k_{z}} \bigg) \sum_{u=0}^{z-1} \Delta_{t-1}^{u} + \gamma_{z}^{k_{z}} \Delta_{t-1}^{z} = \sum_{u=0}^{Z-1} \sum_{z=u+1}^{Z} \bigg(\gamma_{z}^{k_{z}} - \gamma_{z-1}^{k_{z}} \bigg) \Delta_{t-1}^{u} + \sum_{z=1}^{Z} \gamma_{z}^{k_{z}} \Delta_{t-1}^{z}\nonumber\\
= \sum_{u=0}^{Z-1}\Delta_{t-1}^{u}\bigg(\sum_{z=u+1}^{Z} \gamma_{z}^{k_{z}} - \sum_{z=u}^{Z-1} \gamma_{z}^{k_{z+1}} \bigg)+ \sum_{z=1}^{Z} \gamma_{z}^{k_{z}} \Delta_{t-1}^{z} \nonumber\\
= \sum_{u=0}^{Z-1}\Delta_{t-1}^{u}\bigg(\sum_{z=u+1}^{Z-1} (\gamma_{z}^{k_{z}} - \gamma_{z}^{k_{z+1}}) + (\gamma_{Z}^{k_{z}} - \gamma_{u}^{k_{u+1}})  \bigg)+ \sum_{z=1}^{Z} \gamma_{z}^{k_{z}} \Delta_{t-1}^{z} \nonumber\\
= \sum_{u=0}^{Z-1}\sum_{z=u+1}^{Z-1}(\gamma_{z}^{k_{z}} - \gamma_{z}^{k_{z+1}})\Delta_{t-1}^{u} + \gamma_{Z}^{k_{z}} \sum_{z=0}^{Z} \Delta_{t-1}^{z} + \sum_{z=0}^{Z-1} (\gamma_{z}^{k_{z}} - \gamma_{z}^{k_{z+1}}) \Delta_{t-1}^{z} \nonumber\\
= \sum_{u=0}^{Z-1}\sum_{z=u}^{Z-1}(\gamma_{z}^{k_{z}} - \gamma_{z}^{k_{z+1}})\Delta_{t-1}^{u} + \gamma_{Z}^{k_{z}}\sum_{z=0}^{Z} \Delta_{t-1}^{z} \nonumber\\
= \sum_{z=0}^{Z-1}(\gamma_{z}^{k_{z}} - \gamma_{z}^{k_{z+1}})\sum_{u=0}^{z}\Delta_{t-1}^{u} + \gamma_{Z}^{k_{z}}\sum_{z=0}^{Z} \Delta_{t-1}^{z}\\
\shortintertext{Ultimately, we get:}\nonumber\\
\sum_{z=0}^{Z} \Delta_{t}^{z} \leq \underbrace{\epsilon \bigg( \frac{1-\gamma^{k}}{1 - \gamma} \bigg)+ \epsilon \bigg(\sum_{z=0}^{Z-1} \frac{\gamma_{z}^{k_{z+1}}-\gamma^{k_{z}}_{z}}{1-\gamma_{z}} \bigg)}_{\text{variance reduction}}+ \underbrace{\sum_{z=0}^{Z-1}\bigg( \gamma^{k_{z}}_{z} - \gamma^{k_{z+1}}_{z} \bigg)\sum_{u=0}^{z}\Delta_{t-1}^{u} + \gamma^{k}\sum_{z=0}^{Z}\Delta_{t-1}^{z}}_{\text{bias introduction}}
\end{align}
\end{proof}

\end{document}